\documentclass[Afour,sageh,times]{IEEEtran}
\usepackage{moreverb,url}
\usepackage{gensymb}
\usepackage[colorlinks,bookmarksopen,bookmarksnumbered,citecolor= blue,urlcolor=red]{hyperref}
\usepackage{graphicx}
\usepackage{cite}
\usepackage{picinpar}
\usepackage{amsmath}
\usepackage{url}
\usepackage{flushend}
\usepackage[latin1]{inputenc}
\usepackage{colortbl}
\usepackage{soul}
\usepackage{multirow}
\usepackage{pifont}
\usepackage{color}
\usepackage{alltt}
\usepackage{enumerate}
\usepackage{siunitx}
\usepackage{breakurl}
\usepackage{epstopdf}
\usepackage{pbox}
\usepackage{subfigure}
\usepackage{mathrsfs}
\usepackage{algorithm}
\usepackage{algorithmic}
\usepackage{verbatim}
\usepackage{booktabs} 
\usepackage{amsfonts}
\usepackage{balance}
\usepackage{gensymb}
\usepackage{tikz}
\usepackage{verbatim}
\usetikzlibrary{trees}
\usetikzlibrary{decorations.pathreplacing}
\usepackage{threeparttable}
\usepackage{footnote}
\usepackage{forest}
\forestset{
  forest scheme/.style={
    for tree={
      inner sep=0pt,
      outer sep=0pt,
      fit=band,
      child anchor=west,
      parent anchor=east,
      grow'=0,
      anchor=west,
      align=left,
      if n=1{
        edge path'={(!u1.west) -- (!ul.west);}
      }{no edge},
      edge={decorate, decoration={brace,amplitude=1.5mm,mirror,raise=2mm}},
    },
  }
}

\newcommand\BibTeX{{\rmfamily B\kern-.05em \textsc{i\kern-.025em b}\kern-.08em
T\kern-.1667em\lower.7ex\hbox{E}\kern-.125emX}}

\begin{document}


\title{Kaiwu: A Multimodal Manipulation Dataset and Framework for Robot Learning and Human-Robot Interaction}
\author{Shuo Jiang,~\IEEEmembership{}  
        Haonan Li,~\IEEEmembership{}
        Ruochen Ren,~\IEEEmembership{}
        Yanmin Zhou,~\IEEEmembership{}
        Zhipeng Wang*,~\IEEEmembership{}
        Bin He*~\IEEEmembership{}
}

\markboth{IEEE Robotics and Automation Letters}%
{Jiang \MakeLowercase{\textit{et al.}}: XXXX}

\maketitle
\begin{abstract}
Cutting-edge robot learning techniques including foundation models and imitation learning from humans all pose huge demands on large-scale and high-quality datasets which constitute one of the bottleneck in the general intelligent robot fields. This paper presents the Kaiwu  multimodal dataset to address the missing real-world synchronized multimodal data problems in the sophisticated assembling scenario, especially with dynamics information and its fine-grained labelling. The dataset first provides an integration of human, environment and robot data collection framework with 20 subjects and 30 interaction objects resulting in totally 11,664 instances of integrated actions. For each of the demonstration, hand motions, operation pressures, sounds of the assembling process, multi-view videos, high-precision motion capture information, eye gaze with first-person videos, electromyography signals are all recorded. Fine-grained multi-level annotation based on absolute timestamp, and semantic segmentation labelling are performed. Kaiwu dataset aims to facilitate robot learning, dexterous manipulation, human intention investigation and human-robot collaboration research. 

\end{abstract}


\section{Introduction}

Robots represent a key platform for artificial intelligence (AI) and are often considered the primary carrier of Embodied AI \cite{zador2023catalyzing}. With the shift in demand from cyberspace to physical environments, traditional industrial robots are inadequate to adapt to changing environments. There is a growing need for intelligent robots capable of interacting with humans in uncluttered environment \cite{ding2018tri} and master human-level skills. However, the path to attaining human-level skills remains challenging and is still an ongoing research problem.

Datasets are the fundamental part of robots and embodied AI, and cutting-edge methods, including imitation learning or foundation models, all pose significant and growing demands on the high-quality, large-scale and multi-modal datasets \cite{firoozi2023foundation}. Imitation learning is primarily achieved through human demonstration data, and most of the methods are based on vision data \cite{xu2022robot} or combined with depth data. New challenges on the data arise from changes in target skills, shifting from simple, fixed motions required for specific tasks, such as picking up an object, to more complex motions. In addition, demonstration via teleoperation represents another alternative approach. For example, Finn et al. proposed mobile ALOHA \cite{fu2024mobile} which used whole-body teleoperation data collection methods. Tesla proposed to use real-human motion capture data to train the Optimus humanoid robots \cite{zhang2024whole}. To be noteworthy, with the emergence of foundation models \cite{xiao2023robot} including ChatGPT, the importance of data volume and quality in determining the algorithm's upper limits has become more widely recognized and there are a few pioneer works exploring the real-world datasets towards the foundation models for robotics. 

RT Series \cite{brohan2022rt}, \cite{brohan2023rt}, \cite{belkhale2024rt} focus on long-term task samples that include multiple activity instances, emphasizing autonomous environmental perception and logical reasoning ability. The emerging challenges for foundational models now entail a shift from managing relatively short-term tasks to attaining long-term autonomy. Philosophically, long-term autonomy requires interaction with the environment and its context, because the brain-inspired algorithms alone are insufficient. Consequently, embodied AI addresses the interplay between humans, robots, and their environment. 


Nevertheless, current datasets all suffer from the following critical limitations. First, most of the data mainly relies on the videos and thus lacks dynamics information. Computer vision methods in the form of images or videos, although advancing and widely deployed, can only represent kinematics information including trajectory and velocity. However, robot learning is a process of dynamics with the surroundings. Missing dynamics information including force will deteriorates the learning performance, resulting in the superficial learning. Second, there is a lack of a universal, sophisticated, intuitive human-level perception framework. The working environment for intelligent robots changes from fixed and clustered settings including manufacturing scenarios to open and complex settings including in situ homes. The environment not only involves static and dynamic objects but also includes human beings. Previous datasets have only incorporated certain sensing modalities, such as video, Inertial Measurement Unit (IMU), etc. which are inadequate to address the demands of increasingly complex tasks. Furthermore, one reason that the manipulation skills of robots cannot surpass those of humans is the lack of fundamental understanding of neural mechanisms. Additionally, relying solely on video as a modality is insufficient to fully reveal how humans accomplish such complex tasks. 
 
To address the above challenges, this paper presents a multimodal framework and dataset for robot learning and evaluation. The key differences from previous datasets include the integration of multimodal data encompassing humans, robots, and the environment. This dataset not only supports robot learning from human demonstrations but also facilitates the prediction of human intentions to enable more effective policy development. We aim to develop skills sets for robot learning and the name Kaiwu is inspired by the ancient Chinese literature \textit{Tiangong Kaiwu} (also known as \textit{The Exploitation of the Works of Nature}), a comprehensive encyclopedia encompassing a wide range of fields, including agriculture and craftsmanship. The contribution of this paper is as follows:

(1) A multimodal data collection framework is proposed, featuring full situational awareness including manipulation dynamics information, human mannipulation neural signals and attention information and multi-view manipulation vision information. This framework aims at complex scenarios including human assembly, towards the universal manipulation ability for embodied AI. 

(2) High-quality, large-scale multimodal data for long-horizon autonomy are collected, utilizing state-of-the-art ground truth techniques for fine-grained and complex manipulation process, holding potential for future benchmark.

(3) The dataset is accompanied by an annotation of spatio-temporal relationships. Rich and fine-grained cross-modal synchronization data annotation are performed, including 298 annotations of regions of interest as personal attention data, 536,467 annotations of closed area elements for image segmentation, 7,197 motion segmentation events for left and right hand dexterity manipulation, 4,467 gesture event annotations, and 4,959 annotations for gesture classification, which significantly enhanced the cross-modal learning and multimodal fusion capabilities and its interpretability.

The rest of the paper is summarized as follows. In Section \uppercase\expandafter{2}-\uppercase\expandafter{4}, research on related datasets, the experimental setup, process design, and the formulation of the target problems are first introduced. In Section \uppercase\expandafter{5}, a detail introduction to the annotation of Kaiwu dataset is presented. Then, the descriptive statistics and directory structure of the dataset are presented, discussed, and introduced by category in Section \uppercase\expandafter{6}. API, project official website and repository, known issues, proposed improvements, and future research directions are discussed in \uppercase\expandafter{7}-\uppercase\expandafter{8} sections. Finally, the conclusion is drawn in Section \uppercase\expandafter{9}.
\begin{figure*}[htb]\centering
	\includegraphics[height=6.0cm]{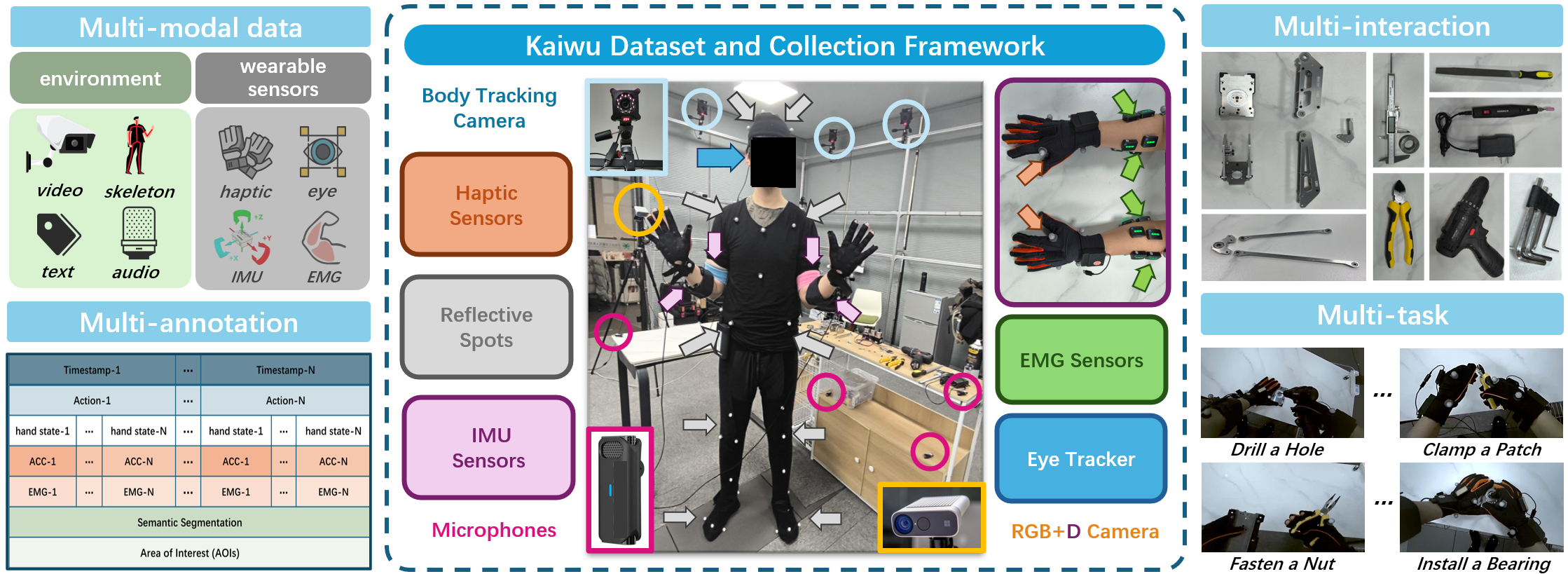}
	\caption{A framework of wearable and environment-mounted sensors records rich activity information in the assembly environment.}
	\label{FIG_all}
\end{figure*}


\section{Related work}

The main focus of this paper is on how to build a multimodal dataset for robot learning via human demonstration for human-level skills and future human-robot collaboration, and thus the related key technology are first reviewed.

\subsection{Robot learning dataset}
Multi-modal input can serve as a crucial foundation for robots to learn human behavior, enhancing the accuracy and efficiency of the learning process.
The dataset focuses on teaching robots to perform daily interactions in changing environments through human demonstrations \cite{xiao2023robot}.


JD ManiData, ManiWAV\cite{liu2024maniwav}, and RH20T \cite{fang2023rh20t} help robots acquire diverse and generalizable skills. The DROID dataset \cite{ khazatsky2024droid} uses a distributed approach to enhance robot performance and generalization. GraspNet-1Billion \cite{fang2023robust} improves object grasping in cluttered scenes with multi-modal data. The Open-X implementation \cite{padalkar2023open} enables large-scale data integration and validates data transfer between robots to improve multi-robot capabilities.

However, the above datasets suffer from data homogenization. The ability to efficiently and purposefully train embodied intelligence models in complex scenarios is limited. Building on the excellent datasets above, more research has been done to further extend on a particular aspect.


The RT series datasets evolve to enhance robot capabilities. RT-1 \cite{brohan2022rt} uses large-scale data and high-capacity models for multitask control in real-world environments, incorporating text, vision, and action data. However, it has limitations in scene and target object selection. RT-2 \cite{brohan2023rt} builds on RT-1 using Web VQA to link images with text tokens, addressing these limitations. The Open X-Embodiment dataset \cite{padalkar2023open} integrates RT-2 with more diverse data to train the versatile RT-X model. The latest RT-H adds human-robot interaction data to further improve learning capabilities.

The ARIO dataset \cite{wang2024all}, based on Open X-Embodiment, includes simulation and real-world data with diverse robot hardware and a uniform format. However, it lacks dynamic kinematic information and authenticity in raw data, and faces labeling generality issues.


\subsection{Human activity recognition dataset}

Enabling robots to recognize and understand human activity is crucial for embodied intelligence. High-quality datasets are essential. The industry-oriented dataset \cite{maurice2019human} captures human movement in industrial settings for classification and prediction. ActionSense \cite{delpreto2022actionsense} focuses on kitchen scenes with multi-modal data. HUMBI \cite{yoon2021humbi} provides a multiview camera dataset for high-resolution human body modeling, enhancing reconstruction and recognition capabilities. The Toyota Smarthome Untrimmed (TSU) dataset \cite{dai2022toyota} collects untrimmed home videos to help robots understand the causality of human activity. Another data set \cite{reily2022real} aims to help robots understand the intention of the team to better assist humans.

The current research lacks an extension of the application scenarios for the dataset and a focus on assembly task-related scenarios. In addition, there is a lack of balance in data modality homogeneity, cross-modality temporal consistency, and task causality analysis.

\subsection{Human robot collaboration dataset}

Embodied intelligence focuses on how intelligent systems interact with their physical bodies and environments. Human-robot collaboration helps agents adapt to surroundings by learning from human interactions, improving the generalizability and stability of their behavior. Multimodal datasets, which include text, images, sound, and video, provide rich information about user intentions and emotions. In human-robot interaction, these datasets enhance agents' understanding of user needs, improve accessibility and perception, and enable more natural and intelligent interactions.


The dataset \cite{huang2019dataset} teaches robots daily interactions in changing environments via human demonstrations. The RBO dataset \cite{martin2019rbo} records human interactions with articulated objects. The RT-H \cite{belkhale2024rt} method enhances RT-2 by adding human-robot interaction interventions, using fine-grained phrases to describe motions. This improves the robot's understanding of interaction actions and language, builds an action hierarchy, and enables learning from human language interventions.


Eye-tracking devices can map the region of interest along the viewing trajectory, helping robots understand human intentions during tasks like assembly. EMG and IMU signals capture muscle behavior patterns during actions, enabling robots to learn and predict hand gestures and respond more accurately to human behavior.


The HARMONIC dataset \cite{newman2022harmonic} captures human-robot interactions in an autonomous setting, using a joystick-controlled robot and wearable sensors to gather data rich in human intent. The HBOD dataset \cite{kang2023hbod} employs more wearable sensors to capture detailed human movements and interactions with tools, enhancing robot understanding of human intentions and maneuvers. The dataset \cite{lastrico2024effects}explores learning challenges in hand-object interactions, including intention recognition and motion generation. The OAKINK2 dataset \cite{zhan2024oakink2} provides multi-view images and accurate pose annotations of humans and objects, supporting applications like interaction reconstruction and motion synthesis.

Regarding some shortcomings of the above datasets,the Kaiwu dataset directly collects dynamic and static data using wearable sensors. It integrates assembly actions into a coherent process, enhancing human-robot interaction with a narrative and causal structure. Additionally, a coordinate framework is established to enrich spatial integrity. A detailed comparison with other state-of-the-art datasets is provided in Table \ref{compare}.

 \begin{table}[htbp]
\renewcommand{\arraystretch}{1.5}
\centering
\caption{\label{compare}A Comparison of related datasets. Motion capture contains 3D skeleton and ground truth.}
\scalebox{0.8}{
\begin{tabular}{lll}
\hline
\textbf{Dataset  }     & \textbf{Modalities}      & \textbf{Environment/Activities}      \\ \hline 
TSU                    & RGB,Depth,3D Skeleton           & Daily actions  \\
Harmonic                 & Gaze,EMG,RGB,Depth             & Meal             \\
Hbod & 3D Skeleton,Tactile,Hand Pose,IMUs     & Tool Operation           \\
Humbi      & RGB,Depth, 3D Skeleton             & Body Expression      \\  
OXE               & Mainly RGB,Depth    & Multiple Scenarios           \\
Actionsense        & IMUs,3D Skeleton,Hand Pose,Gaze& Kitchen Activities \\ &EMG,Tactile,RGB,Depth,Audio    \\
Kaiwu               & IMUs,Motion Capture,Hand Pose,Arm,Gaze& Industrial Assembly \\ &EMG,Tactile,RGB,Depth,Audio              \\  \hline 
\end{tabular}
}

\end{table}



\section{Data collection platform setups}
The data collection platform enables synchronized streaming, storage, and visualization of information from wearable and environment-mounted sensors, accommodating different sampling rates and data formats. Its modular structure simplifies the integration of new sensors. Multi-threading and multi-processing ensure efficient CPU and RAM usage. The platform also supports post-processing, allowing integration with third-party applications. It aims to enhance data collection efficiency by reducing experimental artifacts, enabling easy metadata recording, and allowing researchers to focus more on the subject build process.

\subsection{Application and use cases}
The main purpose of this dataset is to provide rich information on how human can achieve dexterous operation and how these information can help robot acquire human-level intelligence. Therefore, we identify several critical processes of human dexterity including dynamics information during the manipulation, attention during the cognition, understanding of muscle group mechanism and assembly logic, ubiquitous environment information along with the operation process and also ground truth recording. The overview diagram is shown in Fig. \ref{FIG_all}.

\subsection{Sensor setups}
Targeted at the above requirements, cutting-edge sensing acquisition equipment are utilized. These devices include wearable sensors and environment-mounted sensors (Fig. \ref{FIG_all}).

\subsubsection{Data glove}
 \begin{table}[htbp]
\renewcommand{\arraystretch}{1.4}
\centering
\caption{\label{gloveformat}Data format of glove data.}
\scalebox{0.8}{
\begin{tabular}{lll}
\hline

sequence       & column 1           & $sequence$  \\
timestamp        & column 2        & $timestamp$             \\
hand & column 3-6   & $[hand_w, hand_x, hand_y,hand_z]$          \\
forearm      & column 7-10       & $[fore_w, fore_x, fore_y, fore_z]$
      \\  
upper arm      & column 11-14  & $[up_w, up_x, up_y, up_z]$
           \\
angle sensor     & column 15-33 & $[thumb, thindex, index, inmid, middle,$ \\ 
 & & $midring, ring, ringlittle, little, thumbcmc]$\\
tactile sensor       & column 34-52 & $[p0-p18]$ \\  \hline 
\end{tabular}
}

\end{table}

\begin{figure}[htb]\centering
	\includegraphics[height=4cm]{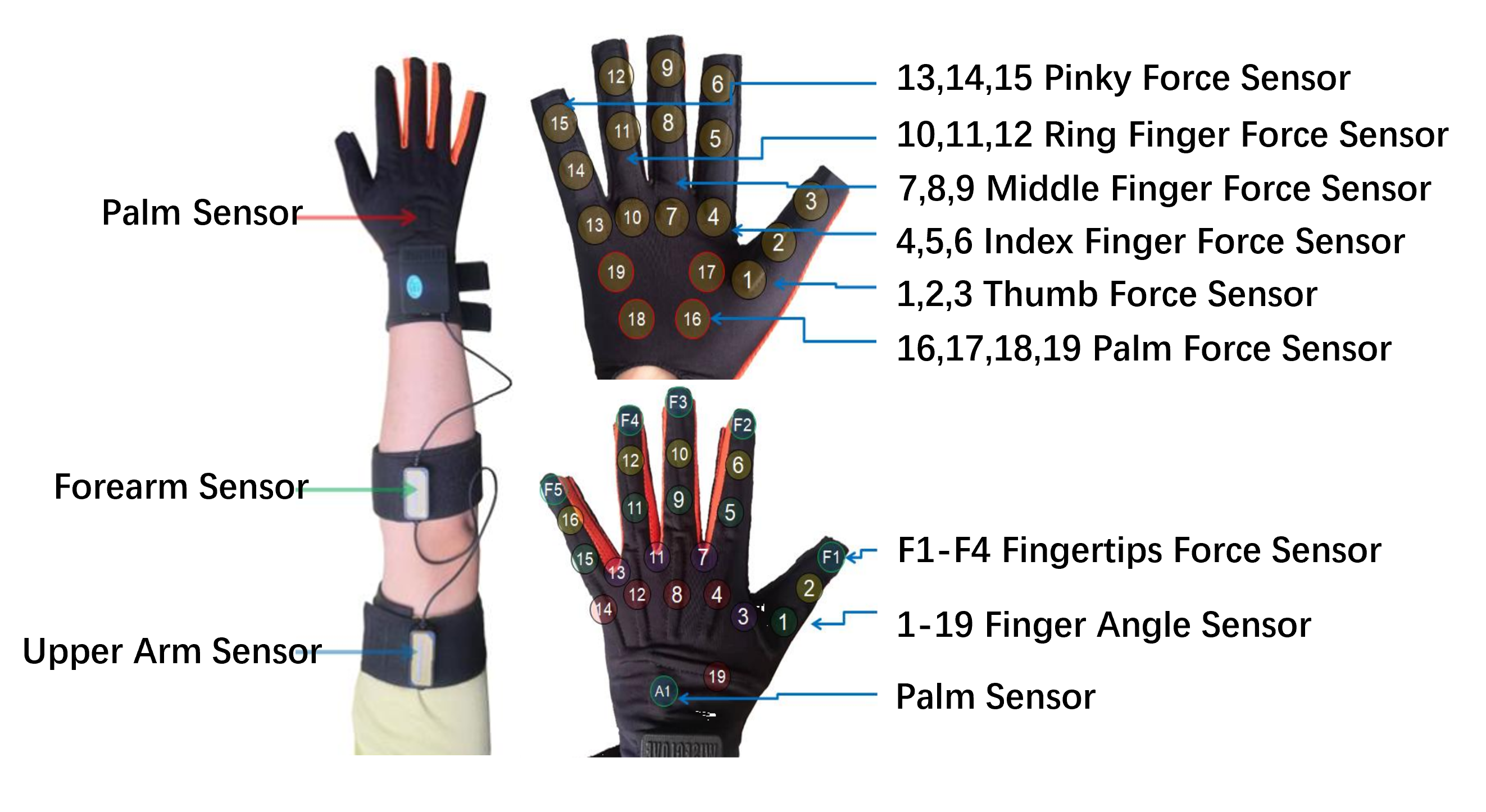}
	\caption{Data glove with angle sensors and force sensors.}
	\label{glove3}
\end{figure}

Data gloves are utilized to simultaneous capture the hand movement and palm tactile interaction information. Therefore, the glove (Fig. \ref{glove3}) is chosen which can drive 3D animated human hand movements in real time. 19 finger angle sensors and 19 finger pressure sensors are equipped with a sensing accuracy of 9g, which fulfills the spatial and accuracy resolution of the experimental requirements. Additionally, there is an arm sensor that recorded quaternion data from the palm, forearm, and upper arm sensors. The system can capture human hand motion in real-time and drives 3D animation simultaneously to depict hand motions. This system is used to study participants' hand movements and finger coordination during assembling tasks.

\subsubsection{EMG and ACC}
\begin{table}[b]
\renewcommand{\arraystretch}{1.4}
\centering
\caption{\label{EMGformat}Data format of EMG (or ACC) data.}
\scalebox{0.94}{
\begin{tabular}{lll}
\hline

index sequence       & row 1           & \textit{data index} \\
timestamp        & row 2        & \textit{timestamp of data}     \\
left hand & row 3-10   & \textit{raw data of EMG or ACC sensors}\\
right hand& row 11-17&\textit{raw data of EMG or ACC sensors}\\  
 \hline 
\end{tabular}
}

\end{table}
The muscle activity of the participants is accurately collected for studying the relevance of EMG data during assembly tasks. The Trigno Sensors are equipped with an integrated 9-degree-of-freedom IMU capable of transmitting data pertaining to acceleration (ACC). This information can be used to discern movement activity that is time-synchronized with the EMG signals. They monitor the electrical signals of the muscles through electrodes attached to the participant's skin and converts them into readable digital data. A total of 16 EMG sensors are used, with 8 sensors attached to the left side and 8 to the right side, each being attached to the basic muscle groups of the forearm.

\subsubsection{Environment depth and visual information}
Camera is positioned directly in front of the participant to record the environmental information including RGB and depth information during the assembly process. This camera is used to capture participants' body postures and movements in real-time for investigating their motor performance and movement control in specific tasks.
\subsubsection{Eye tracker}
Eye-tracking data, including participants' gaze points mobility, are obtained to invesitgate visual attention and cognitive processes. The eye tracker is a binocular stereoscopic dark pupil tracking device that relies on corneal reflexes to record and analyze human eye tracking and first-person information in assembly processes. With an average accuracy of 0.6\degree, an average precision of 0.03\degree and an average data loss rate of 0.01\%, the eye tracker device guarantees quality data collection.
\subsubsection{Environment sound information}
The microphones are used to capture ambient sound as a complement to modality, such as the sound of a tool being placed and rubbing against parts. In the experiment, four microphones are placed in different locations. Numbered 1-2, are placed on the operator's table to record the sound information of the operating environment. Microphones numbered 3-4 are placed in the accessory area to record ambient sound information for gripping tools and parts.

\subsubsection{Ground truth}
The 3D Motion Capture System is used to capture and record the movement and actions of a person or object in three-dimensional space. It utilizes high-precision optical sensors and cameras to track the movement trajectory of the object being measured and then transmitting the data to a computer for analysis and rendering. In the experiment, participants are fitted with reflective markers at 37 key points on the body to calibrate and build a mannequin. With this system, the ground truth information for analysis can be accurately captured. 


\begin{figure*}[t]\centering
	\includegraphics[height=5.5cm]{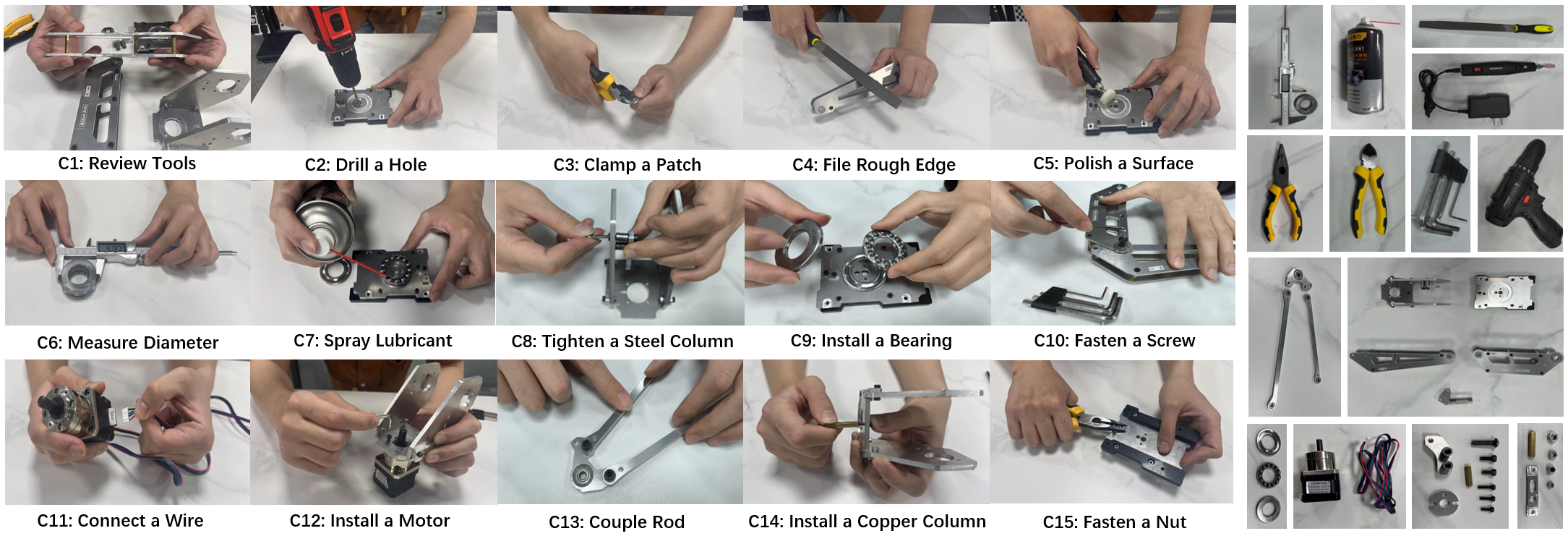}
 	\caption{Overview of assembling process.\textit{ C8} and \textit{C14} have different focuses. The former collected experimental data on columnar parts that participants tightened by hand in a situation where there is ample mounting space. The latter captures experimental data by tools in a confined space with limited hand movement.}
	\label{assembly activity}
\end{figure*}

\subsection{Calibration}

Initial calibration of different participants can significantly improve the quality of metadata.
\subsubsection{Equipment initialization calibration}
The eye tracker uses a one-point calibration procedure to account for individual differences in eye shape and geometry, enhancing accuracy in predicting visual positions. Before experiments, the eye tracker is calibrated for each subject. The data recorder positions a calibration card at arm's length from the subject, who wears the eye tracker's head-mounted module and focuses on the card's center during calibration.

Physical stature varies across participants, so for each collection experiment, the motion capture system needs to be calibrated to the skeleton of the participants after the reflective markers are installed. The participant's arms are opened at shoulder level and slowly rotated to ensure that all reflective markers on the participant's body are recorded in the system. 


When performing arm activity manipulations, the active muscle groups in the lower arm vary from participant to participant. Therefore, it is necessary for the participant to continually rotate the wrist and measure the most significantly active muscle groups to place the EMG acquisition unit to confirm that the most distinctively characterized EMG signal data is captured.

\subsubsection{Process initialization calibration}
After starting the experiments, the participants perform calibration of gesture movements to activate data synchronization and initialize the 19 sensors on the gloves. Each group of experiments can only be performed after completing the calibration process mentioned above for the corresponding assembly operation to ensure the synchronization of data acquisition.
\subsubsection{Data synchronization}

The data collection platform with modular programming and ultimately implemented multi-threading is designed to ensure simultaneous start and synchronized acquisition of each acquisition module. The platform ensures the independence of data collection and the convenience of code adjustment. For different devices with different sampling rate in Subsection 6.1. Absolute timestamp records are used during data collection, and the output of each module's data stream is synchronized through absolute timestamps.
\section{Data collection protocol}

\subsection{Participants}

Twenty participants with a mean height of 172.4$\pm$6.5 cm and a mean age of 23.95$\pm$5.1 yrs are recruited in this process, and all the consents are given. Each volunteer is recorded for 2 hours. The entire data collection process span 11 working days. Approval of all ethical and experimental procedures and protocols is granted by Institutional Review Board.

\subsection{Assembling process}
After the participant completed donning the device, calibration session, and data stream synchronization, data acquisition of the assembly actions began sequentially. During a single action acquisition session, the process would last 140 seconds, involving 0-10 seconds of data synchronization to receive calibration and 130-140 seconds of buffer time.
\subsubsection{Assembly activity}

In the assembly process of the robotic arm, links that closely reflect human dexterity and dynamics are selected for data collection. The entire assembly process is divided into 15 assembly links, with each link divided into left and right fine-grain motion elements labeled as action units. The segmentation  facilitates the coding and integration of the entire assembly process into an action ensemble, and also facilitates subsequent labeling and recall. The assembly activities are shown in Fig. \ref{assembly activity}.

\subsubsection{Assembly equipment and parts}

For the action set, the relevant assembly objects include the robot front arm, end arm, middle arm, arm base, rear arm part, flange bearings, linkage, motor, steering bearings, brass post, steel post, bolt spacer, bolts, nuts, and power cable.

In our experiments, we do not subdivide the action units into finer degrees. Instead, we work with larger components because finger tactile sensors struggle to detect data changes in small, delicate parts. Before data collection, an initial pre-assembly of the small components is performed to meet acquisition requirements.

\subsubsection{Tools} 
For all divided action sets, the following execution tools are required: drill, snips, file, polisher, vernier calipers, sharp-nosed pliers, and screwdriver.



\section{Data annotation}

\begin{figure*}[t]\centering
    \includegraphics[height=5.5cm]{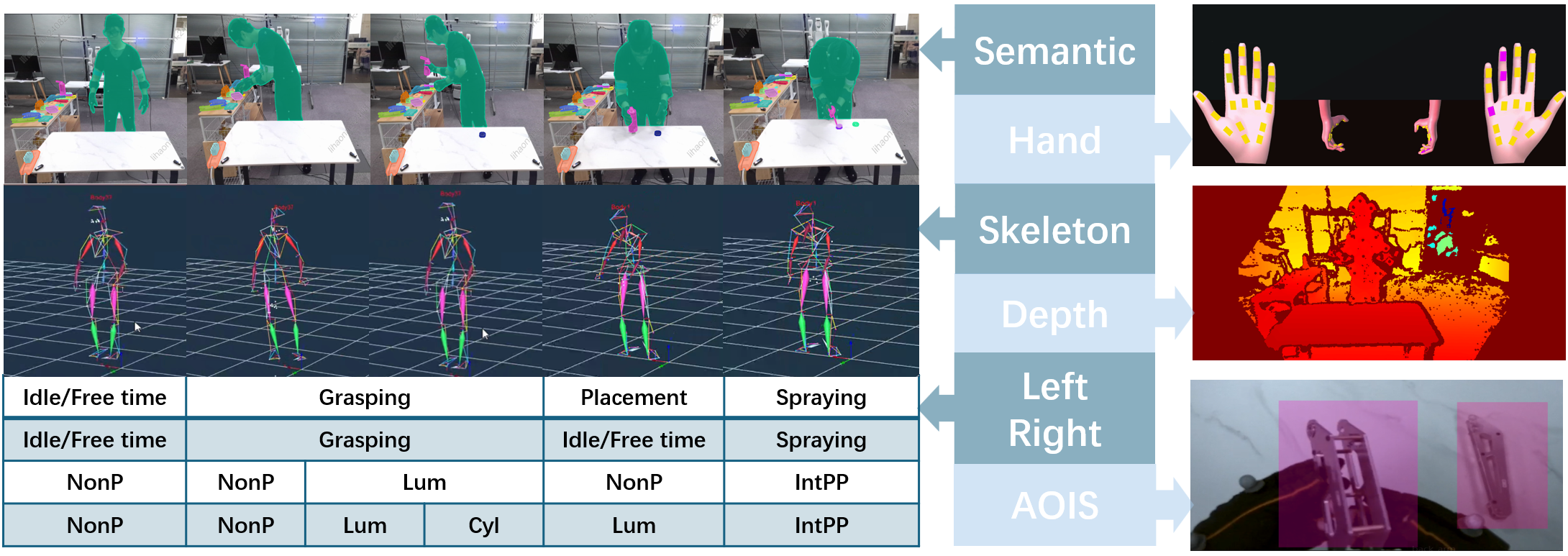}
	\caption{Overview of annotation.}
	\label{annotation review}
\end{figure*}

An initial labeling of the dataset is performed to facilitate potential robot learning and human-robot interaction tasks. Various data are labeled to enhance the usability of the dataset. Subsequent users have the flexibility to annotate data according to their specific project requirements, while our initial annotations serve as a fundamental reference paradigm (Fig. \ref{annotation review}).
The annotation information, such as labels and instance, is shown in Table \ref{sum}.
 \begin{table}[htbp]
\renewcommand{\arraystretch}{1.4}
\centering
\caption{\label{sum}Overview of annotation.}
\scalebox{0.9}{
\begin{tabular}{cccc}
\hline

\textbf{type of mission} & \textbf{annotation element} & \textbf{object tags} & \textbf{instance} \\ \hline
gesture classification & picture & 10 & 4959     \\
AOIs & video & 30 & 298   \\
semantic segmentation & closed area & 30 & 610778   \\  
action segmentation & video clip & 26 & 7197  \\
gesture segmentation & video clip & 9 & 4467 \\ 
\hline 
\end{tabular}
}
\end{table}
\subsection{Action\&gesture segmentation and export}
Fine-grained segmentation is conducted based on the timestamp series of actions and gestures to identify the start and end timestamps of the multimodal data collected from each assembly task action unit. The annotation includes: 
\begin{itemize}
    \item 
    Action-level Annotation: Coarse segmentation of left and right hand actions.
    \item
    Fine-grain Gesture Annotation: Within the coarsely segmented time interval, a fine-grain segmentation of the left or right hand states is made
\end{itemize}
For action segmentation, the left and right-handed action segments of each participant are divided as a whole into time intervals for the stage actions. And the free gaps between movements are retained to preserve the causality and coherence between movements. 

The execution of actions in each segmentation interval is transitioned by the change of several gesture states, the actions of the left and right hands are further divided into corresponding hand state intervals, which are notated as Fine-grain Gesture Annotation. In the division of hand states, all hand states are divided into 8 gesture states \cite{vergara2014introductory}: Cylindrical grasp (Cyl), Oblique palmar grasp (Obl), Lumbrical grasp (Lum), Intermediate power-precision grasp (IntPP), Pinch grasp (Pinch), Lateral Pinch (LatP), Special pinch (SpP), Non-prehensile grasp (NonP). 

Firstly, according to whether or not the palm is involved, then they are divided into Group1: Cyl, Obl, IntPP and Group2: Pinch, LatP, SpP, Lum. in Group1 is further divided by the state of the thumb and the four fingers, which corresponds to Cyl: the thumb and the four fingers are naturally curved; Obl: the thumb is straight and the four fingers are curved; and the others are stipulated as IntPP. in Group2 is further divided by the number of fingers involved in the The number of fingers involved in the activity is further classified as Group2a: a small number (2-3) of fingers involved, containing Pinch and LatP; Group2b: four fingers involved, containing Lum and SpP.Finally, by the presence or absence of the lateral fingers, it is classified as LatP and SpP with lateral finger involvement, and Pinch and Lum without lateral finger involvement.

In summary, for each time sequence, both action segmentation and gesture segmentation dimensions are included. For the labeled timestamped sequences, two annotation-level timestamped sequences in the glove data, EMG, and ACC signal data are derived to achieve absolute timestamped labeling of the EMG-ACC-Glove data series.

\subsection{Semantic segmentation}

Semantic information annotation of 30 key objects in RGB video data is performed. By annotating the camera environment information, which records the third-person view of the participants, the data set is enriched with semantic information following the assembly logic. This enhancement aims to improve the robot's logical understanding of the assembly process and human action learning. 

In semantic segmentation, semantic information is labeled by pumping the captured RGB video at one-second intervals to obtain frames with corresponding timestamps. The participants and all interaction tools are labeled. For objects with multiple entities, each object is labeled followed by the corresponding index for differentiation. For objects that are partially blocked in use, the remaining portion is locally labeled. All labeling errors are controlled within 8 pixels. 

\subsection{Gesture classification}

According to the model playback, the state of the hand can be categorized \cite{gopura2018hand} according to whether the thumb and four fingers are flexed or extended. Compared to the gesture segmentation in subsection 5.1, this part of the state segmentation simplifies the further division of the four-finger state in subsection 5.1.

\subsection{Area of interest (AOIs)}
The exported first-person view videos are labeled with a region of interest based on key object manipulation. This region contains the focus and intentional expression information within the participant's field of view, giving more weight to the data of objects in that specific area.

\section{Data format}

\subsection{Descriptive statistics}
A total of $6\times15\times20$ sets of experimental data on the assembling process are collected for Kaiwu dataset. The data comes from wearable sensors worn by 20 participants. Each participant is asked to complete 15 typical actions specified in the assembly process. Each participant spend approximately 19 minutes, resulting in the dataset representing about 6.3 hours of the assembling process. A summary of the related data is presented in Table \ref{tab:widgets2}.

\begin{table}[b]
\renewcommand{\arraystretch}{1.6}
\centering
\caption{\label{tab:widgets2}Descriptive Statistics.
}
\scalebox{0.85}{
\begin{tabular}{cccccccc}
\hline
\textbf{ Data Type}   & \textbf{Documentation Space}     & \textbf{Sampling Rate}      \\ \hline 
\textbf{Glove Data          }      & 264 MB     & 100 Hz  \\
\textbf{Glove Export        }       & 1,124 MB             & 20 Hz             \\
\textbf{Eye Tracking        }     &    14 GB         & 25 Hz                        \\
\textbf{RGB-D Video         }  & 3,476 GB             & 60 Hz            \\
\textbf{Motion Capture Data }   &4,160 MB       & 60 Hz       \\ 
\textbf{Audio Data          }          &  7,955 MB            & 50 Hz            \\
\textbf{ACC Data            }      & 354 MB          & 40 Hz           \\
\textbf{EMG Data            }      & 362 MB             & 40 Hz           \\
\hline
\end{tabular}
}

\end{table}
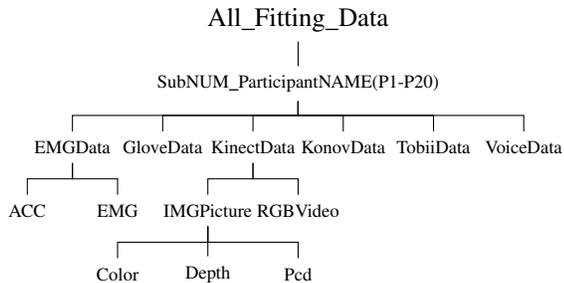
\begin{figure}\centering
\begin{tikzpicture}[scale=0.8, level distance=3em, 
    T/.style={edge from parent fork=#1, grow=#1, level distance=3em, sibling distance=0.5em, +}]
    \node{All\_Fitting\_Data}[edge from parent fork down]
    child {node {\scriptsize SubNUM\_ParticipantNAME(P1-P20)} 
    child {node {\scriptsize EMGData}
    child{node{\scriptsize ACC}}
    child{node{\scriptsize EMG}}
    }
    child {node {\scriptsize GloveData}}
    child {node {\scriptsize KinectData}
    child{node{\scriptsize IMGPicture}
    child{node{\scriptsize Color}}
    child{node{\scriptsize Depth}}
    child{node{\scriptsize Pcd}}
    }
    child{node{\scriptsize RGBVideo}}
    }
    child {node {\scriptsize KonovData}}
    child {node {\scriptsize TobiiData}}
    child {node {\scriptsize VoiceData}}
    };
\end{tikzpicture} 
\caption{KAIWU Dataset Structure Catalog}
\label{ALL}
\end{figure}
\subsection{Detailed information}
This dataset contains experimental data from 20 participants in 20 folders with subject number. The folder corresponding to each participant contains the collected data from 6 recording devices. The directory structure of the dataset is shown Fig. \ref{ALL}.

\subsubsection{EMG data}

The EMGData folder contains the ACC data output by the built-in IMU sensors and the EMG files measured and output by the EMG sensors. The ACC or EMG data are output by each of the built-in IMUs or EMG sensors respectively in 17 sets of data recorded as TIME.csv files. Each file contains 1 row for the calibration standard value and 2-17 rows for the ACC or EMG signals from the 16 sensors. 

\subsubsection{Glove data}
For the assembly actions of C1-C15, the data gloves are transferred and stored as r/l\_TIME.csv files with columns 1-12 representing the quaternion data of the palm, forearm, and upper arm, columns 13-31 representing the finger angle sensor data, and columns 32-50 representing the grip force sensor values. In addition, the data can be exported to present a visualization interface of the hand activity during data collection, and stored as an MP4 file. 
\subsubsection{RGB-D data}
RGB-D data is stored in the C1-C15.mkv files in the RGB\_video file and contains the timestamps used to generate the picture data, recorded as .txt files. The data generated based on the timestamp is stored in the IMG\_picture file, including RGB images (.jpg file and .pcd file), depth information (.png file). The environmental information of the assembly data of C1-C15 is stored under 15 folders respectively. 
\subsubsection{Ground truth data}
After the initial body joint point calibration, the Motion Capture system captures and records the 3D spatial coordinates of each marker point as the assembly process proceeds, outputting the data as a (.cap .trb .xrb) File. Visualization is achieved by reimporting the data.
\subsubsection{Eye tracker data}
The Tobii-data folder includes first-person perspective data (tobii-data) and eye-tracking data (tobii-export). Tobii-data contains compressed .gz files (eventdata for events, gazedata for eye-tracking, imudata for IMU measurements) and a visualization .mp4 file. Tobii-export includes .mp4 files showing gaze point movements for C1-C15 and .xlsx files with data tables from the gyroscope, magnetometer, accelerometer, and eye tracker. Each file header is labeled with the corresponding data meaning.
\subsubsection{Voice data}
The microphone outputs environment sound information as .wav audio files and .txt timestamp data. Microphones numbered 1-2 record the ambient sound in the assembly area and 3-4 record the ambient sound in the accessory area.
\\

\section{Accessing the data}
The data has been uploaded to ScienceDB with DOI number. The homepage of this dataset can be searched using the keyword "Kaiwu" on ScienceDB. The following files are provided for download: A compilation of all the data \texttt{Kaiwu Data Annotation} (31.84GB), which \texttt{action segmentation} (963MB), contains the timestamp exported \texttt{delsys\_data} with ACC and EMG and the \texttt{gloves\_data} with action labels. The \texttt{AOIs} (17.9GB) contains the \texttt{annotation results} and \texttt{annotation process} files which that can be opened using Tobii Lab software. The \texttt{gesture segmentation} (973MB) contains the \texttt{delsys\_data} and \texttt{gloves\_data} with fine grain gesture labels. The \texttt{semantic segmentation} (11.6GB) contains the \texttt{part1-4.zip} with RGB video annotation results, and finally \texttt{gesture classification} (417MB), contains the hand state classification data(\texttt{classification result.zip}). 

In addition, the dataset contains some missing data and additions to the raw data (\texttt{action segmentation supplement} and \texttt{raw data} directories). The \texttt{action segmentation supplement} (193MB) adds four sets of experimental data that are missing from the action segmentation. The \texttt{raw data} (167GB) contains the raw voice data, motion capture data, RGB-D data, glove exported video and eye tracker exported video.
The annotated data is saved and encapsulated as a .csv file. The timestamps can be used as indexes to call the corresponding data sources of different sensors at the same time, and contain the data annotation information of the corresponding timestamps. 




\section{Conclusion}
This paper introduces a data collection framework and platform to build Kaiwu dataset, which enable to improve understanding of human assembly activity logic and robot learning. Multi-modal wearable sensors are equipped to capture various egocentric data which contains rich visual and dynamics info. This is also coupled with environmental and global ground truth info to build an absolute spatial coordinate frame. An initial labeling of the dataset is also annotated to facilitate the training of deep neural networks with multi-modal data.


Future work can utilize the Kaiwu dataset to develop knowledge pathways for embodied learning and analysis, exploring topics such as cross-modal prediction, assembly logic sequence prediction, assembly task planning, and robot self-assembly. In addition, we hope the collection platform of the Kaiwu dataset can be utilized as a medium for robot behavioral learning and human-robot skill transfer. In addition, more advanced wearable sensors can be added to explore new collection modes flexibly to enable the dataset for various application scenarios, serving as the benchmark for the database of robot foundation models, and thus pave the way for general embodied intelligence.


\bibliographystyle{Bibliography/IEEEtranTIE}
\bibliography{Bibliography/BIB_1x-TIE-2xxx}\ 

\end{document}